\documentclass{article}

\usepackage[authoryear, compress]{natbib}
\usepackage[utf8]{inputenc} % allow utf-8 input
\usepackage[T1]{fontenc}    % use 8-bit T1 fonts
\usepackage{hyperref}       % hyperlinks
\usepackage{url}            % simple URL typesetting
\usepackage{booktabs}       % professional-quality tables
\usepackage{amsfonts}       % blackboard math symbols
\usepackage{nicefrac}       % compact symbols for 1/2, etc.
\usepackage{microtype}      % microtypography
\usepackage{amsmath}        % common math package
\usepackage{bm}             % bold math symbols
\usepackage{algorithm}      % algorithm
\usepackage[noend]{algpseudocode}
\usepackage{graphicx}
\usepackage{xcolor}
\usepackage{multirow}
\usepackage{subcaption}
\usepackage{cleveref}
\graphicspath{ {fig/} }

\DeclareMathOperator*{\argmax}{argmax} 

\title{Linear Discriminant Generative Adversarial Networks}

\author{
  Zhun Sun, Mete Ozay, Takayuki Okatani \\
  Graduate School of Information Sciences\\
  Tohoku University\\
  \texttt{sun,mozay,okatani@vision.is.tohoku.ac.jp} \\
}

\begin{document}

\newcommand\crule[3][black]{\textcolor{#1}{\rule{#2}{#3}}}
% \newcolumntype{K}[1]{>{\centering\let\newline\\\arraybackslash\hspace{0pt}}m{#1}}
\newcommand{\specialcell}[2][c]{%
  \begin{tabular}[#1]{@{}c@{}}#2\end{tabular}}

\maketitle

\begin{abstract}
We develop a novel method for training of GANs for unsupervised and class conditional generation of images, called Linear Discriminant GAN (LD-GAN). The discriminator of an LD-GAN is trained to maximize the linear separability between distributions of hidden representations of generated and targeted samples, while the generator is updated based on the decision hyper-planes computed by performing LDA over the hidden representations. LD-GAN provides a concrete metric of separation capacity for the discriminator, and we experimentally show that it is possible to stabilize the training of LD-GAN simply by calibrating the update frequencies between generators and discriminators in the unsupervised case, without employment of normalization methods and constraints on weights. In the class conditional generation tasks, the proposed method shows improved training stability together with better generalization performance compared to WGAN (\citealt{arjovsky2017wasserstein}) that employs an auxiliary classifier.
\end{abstract}

\section{Introduction}
Generative Adversarial Networks (GANs,~\cite{goodfellow2014generative}) have shown great success in generating authentic images. In a GAN, a generator is trained together with a discriminator simultaneously, by performing an adversarial game in which the discriminator is trained to distinguish real samples from those generated by the generator. Although GANs have demonstrated very impressive progress on various synthesizing tasks (\citealt{ledig2016photo, reed2016generative, mathieu2015deep, zhu2016generative, yan2016attribute2image}), training of such networks is known to be demanding due to the instability caused by vanishing and exploding gradient problems (\citealt{arjovsky2017towards}). In earlier works, several variations of the vanilla GAN have been developed to provide better stability and convergence (\citealt{SalimansGZCRC16, huang2016stacked, 2016arXiv160600709N, metz2016unrolled, poole2016improved}). Recently, a new family of GANs that either explicitly (\citealt{mroueh2017mcgan}) or implicitly (\citealt{arjovsky2017wasserstein, 2017arXiv170106264Q}) employ \textit{moment matching objectives} have achieved impressive progress in stabilizing the training of GANs. Notably, \cite{arjovsky2017wasserstein} proposed the Wasserstein GAN (WGAN) that employs Wasserstein distance instead of Jensen–Shannon (JS) divergence to address the vanishing gradient problems of the vanilla GAN. \cite{2017arXiv170106264Q} suggested a similar approach by using Lipschitz constraints on the data probability densities, and proposed a generalized version of WGAN, namely generalized loss-sensitive GAN. 

Yet another open problem is conditional training of GANs, where additional information such as class labels is considered for generating samples correspondingly. A lot of efforts have been made to develop conditional generation methods for the vanilla GAN (\citealt{mirza2014conditional, springenberg2015unsupervised, SalimansGZCRC16, ChenDHSSA16, odena2016conditional}), and regarding the moment matching GAN family, the most common approach is to adopt an auxiliary classifier (\citealt{mroueh2017mcgan, 2017arXiv170106264Q}). However, \cite{2017arXiv170106264Q} pointed out that this approach will indeed impair the generalization properties of GANs, and a trade-off need to be made to balance between cost functions of classification and generation objectives. 

One of the challenges of training GANs, whose generators are updated by matching the first order moment, is to obtain a discriminator with decent separation capacity between the generated and targeted sample distributions. If we assume that the samples can be represented by Gaussian distributed hidden representations, then the aforementioned problem can be addressed using Linear Discriminant Analysis (LDA), straightforwardly. Motivated by this, in this paper we implement the discriminator with an LDA to provide decision hyper-planes for training of generators. Our main contributions can be summarized as follows:
\begin{enumerate}
    \item We develop a novel method for training of GANs, called Linear Discriminant GAN (LD-GAN), whose discriminator is trained to maximize the linear separability between the distributions of hidden representations of generated and targeted samples, while the generator is updated based on the decision hyper-planes provided by performing LDA over the hidden representations.

    \item We utilize the objective of discriminator as a concrete metric of separation capacity of the discriminator, we propose a \textit {decayed incremental learning} of the discriminator, together with a training scheme that calibrates the update frequencies of generators and discriminators dynamically to further stabilize and accelerate the training of LD-GAN. The experimental results from unsupervised image generation tasks demonstrate that our LD-GAN can be smoothly trained to generate authentic images, without employment of neither normalization methods (\citealt{ioffe2015batch, salimans2016weight}) nor additional constraints (decay or clip) on weights. 

    \item We further expand the LD-GAN to generate images with label conditions, by performing LDA to discriminate real and generated samples as well as class-wise samples simultaneously. The experimental results indicate improved training stability and better generalization performance of the LD-GAN, compared to WGAN that employs an auxiliary classifier.
\end{enumerate} 

\section{Background}
\subsection{Generative Adversarial Networks}
\noindent \textbf{Vanilla GAN:} A GAN is formulated as a two-player game, where the generator $G_\theta$ (parameterized by $\theta$) takes a random seed vector $\bm{z}$ as an input, and produces a sample $G_\theta(\bm{z})$ in the data space, while the discriminator $D_\phi$ (parameterized by $\phi$) identifies whether a certain sample comes from the true data distribution $P_r(\bm{x})$ or the generator. Subsequently, the discriminator will be updated according to the classification performance between generated and real samples, while the generator will obtain gradients from these samples that cannot deceive the discriminator. An objective of a vanilla GAN can be formalized by
\begin{equation}
    V(G,D) = \underset{\theta}{\min}~\underset{\phi}{\max}~\underset{\bm{x}\sim P_r(\bm{x})}{\mathbb{E}}[\log(D_\phi(\bm{x}))] + \underset{\bm{z}\sim P(\bm{z})}{\mathbb{E}}[\log(1 - D_\phi(G_\theta(\bm{z})))],
\end{equation}
where $P(\bm{z})$ is an arbitrary noise distribution such as the uniform distribution or the normal distribution. In this paper, we use $\bm{z} \sim \mathcal{N}(\bm{0}, \bm{1})$ for all experiments. The training of GANs is known to be difficult; one reason is that it is demanding to make balance between updates of the generator and discriminator. Literally, an optimal discriminator $D^*_\phi$ is required to correctly estimate the ratio between generated and real data distribution $P_g(\tilde{\bm{x}})/P_r(\bm{x})$. Thus, one can start minimizing the Jensen-Shannon divergence ($f$-divergence in general) by minimizing the objective function of the generator~(\citealt{goodfellow2014generative, 2016arXiv160600709N}), however Jensen-Shannon divergence causes vanishing gradients as the discriminator saturates~(\citealt{arjovsky2017towards}).

\subsection{GANs with Moment Matching Objectives}
In the recent works, a lot effort has been spent to develop generative models that match the first or second order moments of hidden representations as their objectives. In the early works, generative models with Maximum Mean Discrepancy objective (MMD) training was first proposed by \citealt{li2015generative} and \citealt{dziugaite2015training}. \citealt{SalimansGZCRC16} demonstrated that it is possible to train generator of a GAN by matching the mean feature extracted from the discriminator. \citealt{mroueh2017mcgan} further proposed a GAN that is trained by matching statistics of distributions embedded in a finite dimensional feature space. Besides the GANs that employ moment matching objective explicitly, there exist several types of GANs implementing moment matching in practice, despite that they were proposed through other motivations initially, such as Energy-Based GAN~(\citealt{zhao2016energy}), Loss-Sensitive GAN~(\citealt{2017arXiv170106264Q}) and Wasserstein GAN~(\citealt{arjovsky2017wasserstein}). In this paper, we employ a WGAN as a reference model due to its success in training stability. 

\noindent \textbf{Wasserstein GAN:} The WGAN is proposed to solve the vanishing gradients problem using the Wasserstein distance (also known as earth mover's distance) between distribution of generated samples $P_g(\tilde{\bm{x}})$ and real samples $P_r(\bm{x})$. The discriminator approximates the duality of Wasserstein distance, and the objective can be formalized by
\begin{equation}
    V(G,D) = \underset{\theta}{\min}~\underset{\phi, D \in \mathcal{L}_1}{\max}~\underset{\bm{x}\sim P_r(\bm{x})}{\mathbb{E}}[D_\phi(\bm{x})] - \underset{\bm{z}\sim P(\bm{z})}{\mathbb{E}}[D_\phi(G_\theta(\bm{z}))],
\end{equation}
where $\mathcal{L}_1$ is the set of 1--Lipschitz functions. This objective is shown to be able to provide meaningful gradients for training the generator in proportion to the approximated Wasserstein distance.

In practice, weights of a discriminator are clipped within a compact space $[-c, c]$ in order to provide a Lipschitz continuous function, which limits the capacity of the discriminator. However, since the Wasserstein distance varies whenever the generator is updated, the discriminator needs to be updated constantly in order to provide an approximation of the Wasserstein distance. Furthermore, since the updates of discriminator are mini-batch based and only a few of generated and real samples are considered, it is unstable to employ a momentum based optimizer such as Adam~(\citealt{kingma2014adam}) as the original paper pointed out, and optimization of discriminator is computationally costly. In the implementations, the discriminator is updated four times more than the generator within one iteration.

\subsection{Class Conditional Generation with GANs}
Generative adversarial networks can be extended to conditional models if both the generator and discriminator are conditioned on an additional variable $y$. The variable can be a representation of any type of auxiliary information, such as class labels that represent categories. Various approaches have been proposed for conditional generation with vanilla GANs by either using $y$ as the side information to train the discriminator or tasking the discriminator with reconstructing side information (\citealt{springenberg2015unsupervised, SalimansGZCRC16, odena2016conditional, ChenDHSSA16}). GANs, which perform matching of statistical moments, usually follow the approach that employs an auxiliary classifier to provide calibrated gradients for conditional generation~(\citealt{2017arXiv170106264Q, mroueh2017mcgan}). However, this approach does not only rely on the discriminative capacity of the classifier, which is usually implemented as a cross-entropy loss on the top of its discriminator. But also, a trade-off must be made to balance between classification and generation objectives as pointed out by \cite{2017arXiv170106264Q}.

\subsection{Linear Discriminant Analysis}
\label{sec:LDA}
Linear Discriminant Analysis (LDA) methods are used to compute a linear combination of features which characterize or separate two or more classes of objects. The resulting combination may be used as a linear classifier, or used for dimensionality reduction before employing classification. The transformation is based on maximization of a ratio of ``between-class variance'' to ``within-class variance'' to reduce data variation in the same class, and to increase the separation between classes. 

Formally, let $\mathcal{X} = \{\bm{x}_n \in \mathbb{R}^{M}\}_{n=1}^N $ be a set of $N$ samples belonging to $C$ classes. LDA computes a linear projection $\bm{W} \in \mathbb{R}^{L \times M}$ into a lower dimensional subspace. The resulting linear combinations of features $\bm{XW}^T$, where $\bm{X} = [\bm{x}_1, \bm{x}_2, \ldots, \bm{x}_N ] \in \mathbb{R}^{N \times M}$, are maximally separated in this space (\citealt{bishop:2006:PRML}). The LDA objective used to find an \textit{optimal} projection matrix $\bm{W^*}$ is formulated by
\begin{equation}
    \bm{W^*} = \underset{\bm{W}}{\argmax}~ \frac{\bm{WS_bW^T}}{\bm{WS_wW^T}},
\end{equation}
where $\bm{S_w}$ and $\bm{S_b}$ are the within and between class scatter matrix, which are computed by
\begin{equation}
   \bm{S_w} = \sum_{c \in \mathcal{C}} \sum_{\bm{x} \in \mathcal{X}_c} (\bm{x} - \bm{\mu_c})(\bm{x} - \bm{\mu_c})^T, \;\;\;\;\;\; \bm{S_b} = \sum_{c \in \mathcal{C}} N_c (\bm{\mu_c} - \bm{\mu})(\bm{\mu_c} - \bm{\mu})^T,
\end{equation}
where $C$ is the number of classes, $\mathcal{X}_c \subset \mathcal{X}$ is the set of samples belonging to the $c^{th}$ class, $N_c = |\mathcal{X}_c|$ is the number of samples in the $c^{th}$ class, $\bm{\mu}$ and $\bm{\mu_c}$ are the mean vectors of all samples and samples in the $c^{th}$ class, $\forall c \in \mathcal{C}, \mathcal{C}= \{1,2,\ldots,C\}$.
\vspace{-10pt}
\section{Linear Discriminant Generative Adversarial Networks}
\vspace{-5pt}
Inspired by~\cite{stuhlsatz2012feature} and \cite{dorfer2015deep}, the discriminator of LD-GAN is implemented as an end-to-end combination of an LDA with a feature extractor $R_\phi(\bm{x})$. The discriminator employs an eigenvalue-based objective function that maximizes the linear discrimination between different sources of inputs (e.g. distributions of real and generated samples). Let $\bm{\lambda}$ denote the vector of non-travail eigenvalues of $\bm{S_b}^{\frac{1}{2}} \bm{S_w}^{-1} \bm{S_b}^{\frac{1}{2}T}$, and $\bm{W} = \big[\bm{w}_1, \bm{w}_2, \ldots, \bm{w}_L \big]$ be the matrix of the corresponding eigenvectors. Then the objective function can be reformulated by 
\begin{equation}
\label{eq6}
    V(R) = \underset{\phi}{\max} ~\underset{\bm{x}\sim P_r(\bm{x}), \bm{z}\sim P(\bm{z})}{\mathbb{E}}[\bm{\lambda}],
\end{equation}
where $\mathbb{E}[\cdot]$ is the expectation operator computed over $P_r(\bm{x})$ and $P(\bm{z})$. Intuitively, the objective of feature extractor (discriminator) is to provide discriminative hidden representations for a maximized linear separation between the generated and real samples. Then the generator moves the generated samples towards the provided hyper-plane of the desired data distribution by % using an orthogonal projection
\begin{equation}
    V(G) = \underset{\theta}{min} ~\underset{\bm{x}\sim P_r(\bm{x}), \bm{z}\sim P(\bm{z})}{\mathbb{E}} [\mathcal{H}_r(\bm{u}_g) - \mathcal{H}_g(\bm{u}_g)],
\end{equation}
where $\bm{u}_g = R_\phi(G_\theta(\bm{z})))$ are the hidden representations of samples in an $M$-dimensional space. $\mathcal{H}(\bm{u})$ is the distance of a sample to a linear decision hyper-plane in the $L$-dimensional projected space, which can be computed by
\begin{equation}
\label{eq8}
    \mathcal{H}(\bm{u}) = \bm{u}\bm{\mathcal{A}}^T - \frac{1}{2}diag(\bm{\mathcal{M}}\bm{\mathcal{A}}^T),
\end{equation}
where $\bm{\mathcal{M}} = \big[\bm{\mu}_r, \bm{\mu}_g \big]$ is a matrix of mean vectors of hidden representations of real and generated samples, $\bm{\mathcal{A}} = \bm{\mathcal{M}}\bm{W}\bm{W}^T$ are normal vectors of the linear decision hyper-planes.

Although the dimension of hidden representations $\bm{u}$ can be arbitrary, the rank of $\bm{S}_b$ is $1$ for the unsupervised case where samples are distinguished by ``real'' or ``generated''. Therefore, the projection maps vectors to a $1$-dimensional space, and $\bm{WW}^T$ becomes a scalar. The objective of generator can be simplified to minimize the $l_2$ distance $(\bm{\mu}_r - \bm{\mu}_g)^2$ between the mean vectors. Alternatively, minimizing the eigenvalue $\bm{\lambda}$ can be also employed as the objective. However, since $\bm{W}$ is invariant to scaling, this type of objective will result in an unbounded variance for $\bm{u}_g$ if there are no constraints on weights of discriminator, and quality of generated samples is decreased in practice. 
% which can be reviewed as $-V(R)$ without considering within class variance. 

\begin{algorithm}[t]
\caption{Unsupervised learning algorithm of our proposed LD-GAN with \textit{dynamic balancing}.}
\begin{algorithmic}[1]
\State $\textbf{input:~} \textit{Update scheme}~\mathcal{F}, \textit{decay coefficient}~\eta.$
\For {$\textit{Some training iterations}$}
\State $\textit{Sample a minibatch of noise}~\bm{z}\sim{P(\bm{z})};$
\State $\textit{Generate faked data}~\bm{x_g} = G_\theta(\bm{z});$
\State $\textit{Sample a minibatch of real sample}~\bm{x_r}\sim{P_r(\bm{x})};$
\State $\textit{Extract hidden representations}~\bm{u_g} = R_\phi(\bm{x_g})~and~\bm{u_r} = R_\phi(\bm{x_r});$
\State $\textit{Sample a minibatch of noise}~\bm{z}\sim{P(\bm{z})};$
\State $\textit{Obtain update scheme with}~I_d, I_g = \mathcal{F}(\hat{\bm{\lambda}});$
\For {$I_d~\textit{iterations}$}
\State $\textit{Compute}\{\mathcal{\bm{M}}, N\}~\textit{of the mini-batch and update the incremented}~\{\hat{\mathcal{\bm{M}}}, \hat{N}, \hat{\bm{S}}_b, \hat{\bm{S}}_w\}$
\State $\textit{Compute} \{ \hat{\bm{\lambda}}, \hat{\bm{W}}\}~\textit{and update the feature extractor}~R_\phi~\textit{by descending the gradient of (6)}$
\State $\textit{Decay the}~\hat{N},\hat{\bm{S}}_w~\textit{by multiplying}~\eta$
\EndFor
\For {$I_g~\textit{iterations}$}
\State $\textit{Update the generator}~G_\theta~\textit{by descending the gradient of (7)}$
\EndFor
\EndFor
\end{algorithmic}
\end{algorithm}

\noindent \textbf{Relationship to the Least-Square GAN:}~\cite{2016arXiv161104076M} proposed a GAN that employs least square error between samples and coding as its objective function. It is well-known that the objective of a binary LDA is equivalent to that of the least-square linear regression with coding $\frac{N_r + N_g}{N_r}$ and $-\frac{N_r + N_g}{N_g}$ for real and generated samples~(\citealt{bishop:2006:PRML}), respectively. While the LS-GAN is a non-parametric model that focuses on penalizing individual samples that are away from the given coding in both discriminator and generator, the proposed LD-GAN can be seen as a first order moment matching method with Gaussian assumption of the hidden representations.

% (The computational complexity for invert $bm{S_w}$ and $bm{S_b}$ are $\mathcal{O}(n^3)$)

\noindent \textbf{Incremental learning of discriminators:} As aforementioned, updating a discriminator in a mini-batch style results in an inaccurate approximation of the targeted distribution, and costs more iterations for updating the discriminator in practical implementations. Therefore, an incremental (online) learning of the discriminator is useful to obtain better stability, which could be easily implemented while updating the LDA. We follow the approach proposed by~\cite{pang2005incremental}, and further introduce a decay coefficient for re-weighting the importance of former batches along with the update of the discriminator. That is, in each iteration, we first compute $\{\mathcal{\bm{M}}, N\}$ for a mini-batch and calculate the total $\{\hat{\mathcal{\bm{M}}}, \hat{N}, \hat{\bm{S}}_b, \hat{\bm{S}}_w\}$. Then, we preform an LDA method to obtain $\hat{\bm{\lambda}}$ as the objective. After updating the discriminator, we multiply the $\hat{N},\hat{\bm{S}}_w$ with a $\eta \leq 1$ to decrease the the importance of former batches. The incremental learning of the discriminator allows the generator to be updated more frequently since the obtained decision hyper-planes are more stable, compared to that obtained using only a mini-batch. %thus a faster convergence can be achieved.

\noindent \textbf{Dynamic balancing:} With the parametric assumption, the divergence between hidden representations of generated and real samples can be represented by the eigenvalues $\bm{\lambda}$ straightforwardly. Thus, it is able to balance the update frequency between discriminator and generator during training with a given scheme $\mathcal{F}$, e.g. the discriminator is updated more frequently than the generator when the eigenvalues get smaller, and vice versa. Keep in mind that in \eqref{eq6}, the objective of our proposed discriminator is unbounded, similar to that of the Wasserstein GAN, however we do not employ constraints such as weight decay or weight clipping to bound the mean discrepancy explicitly. In the experimental analyses, we observed that the implicit constraints implemented by dynamic balancing act as a good regularization method and provide a faster convergence compared to employment of explicit constraints.

\begin{figure}[t]
\centering
    \begin{subfigure}{1\textwidth}
    \centering
        \begin{minipage}[c]{0.40\textwidth}
            \includegraphics[width=1\textwidth]{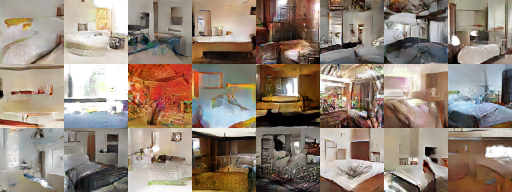}
        \end{minipage}
        \begin{minipage}[c]{0.04\textwidth}
            \crule[white]{0.3cm}{0.3cm}  
        \end{minipage}
        \begin{minipage}[c]{0.40\textwidth}
            \includegraphics[width=1\textwidth]{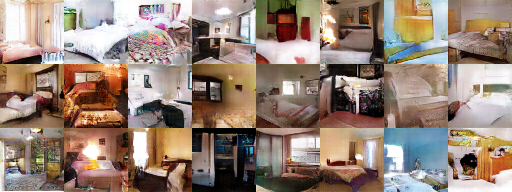}
        \end{minipage} \\
        \begin{minipage}[c]{0.96\textwidth}
            \centering
            \subcaption{Results obtained using weight clip.}
            \vspace{-2pt}
            \label{fig1:a}
        \end{minipage}
    \end{subfigure}
    
    \begin{subfigure}{1\textwidth}
    \centering
        \begin{minipage}[c]{0.40\textwidth}
            \includegraphics[width=1\textwidth]{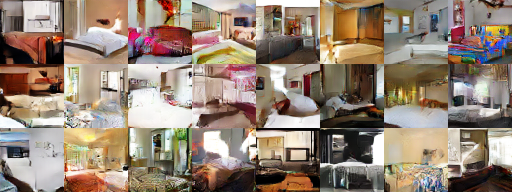}
        \end{minipage}
        \begin{minipage}[c]{0.04\textwidth}
            \crule[white]{0.3cm}{0.3cm}  
        \end{minipage}
        \begin{minipage}[c]{0.40\textwidth}
            \includegraphics[width=1\textwidth]{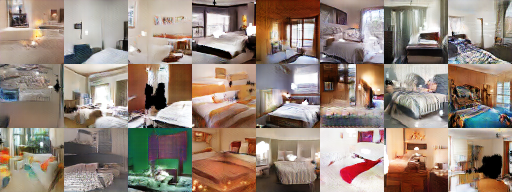}
        \end{minipage} \\
        \begin{minipage}[c]{0.96\textwidth}
            \centering
            \subcaption{Results obtained without using weight clip.}
            \vspace{-2pt}
            \label{fig1:b}
        \end{minipage}
    \end{subfigure}
    
    \begin{subfigure}{1\textwidth}
    \centering
        \begin{minipage}[c]{0.40\textwidth}
            \includegraphics[width=1\textwidth]{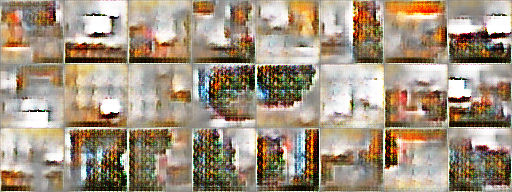}
        \end{minipage}
        \begin{minipage}[c]{0.04\textwidth}
            \crule[white]{0.3cm}{0.3cm}  
        \end{minipage}
        \begin{minipage}[c]{0.40\textwidth}
            \includegraphics[width=1\textwidth]{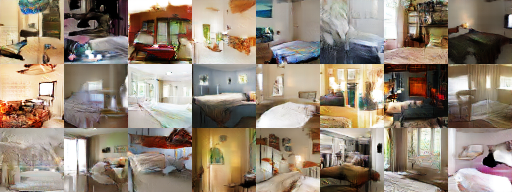}
        \end{minipage} \\
        \begin{minipage}[c]{0.96\textwidth}
            \centering
            \subcaption{Results obtained without using BN in both G and D.}
            \vspace{-2pt}
            \label{fig1:c}
        \end{minipage}
    \end{subfigure}
    
    \begin{subfigure}{1\textwidth}
        \centering
        \begin{minipage}[c]{0.40\textwidth}
            \includegraphics[width=1\textwidth]{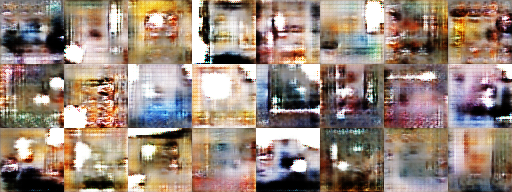}
        \end{minipage}
        \begin{minipage}[c]{0.04\textwidth}
            \crule[white]{0.3cm}{0.3cm}  
        \end{minipage}
        \begin{minipage}[c]{0.40\textwidth}
            \includegraphics[width=1\textwidth]{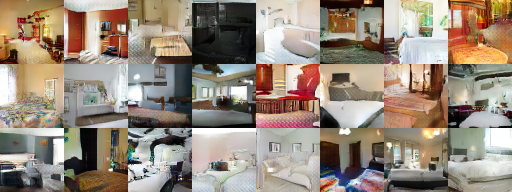}
        \end{minipage} \\
        \begin{minipage}[c]{0.96\textwidth}
            \centering
            \subcaption{Results obtained by optimizing models with Adam.}
            \vspace{-2pt}
            \label{fig1:d}
        \end{minipage}
    \end{subfigure}
    
    \caption{Unsupervised generation using different configurations. \textit{left}: WGAN, \textit{right}: LD-GAN}
    \label{fig1}
\end{figure}

\subsection{Conditional generation with LD-GAN}
\label{subsec_condition}
In order to implement conditional generation in LD-GAN, the objective of the feature extractor is determined to provide discriminative hidden representations for a maximized linear separation between $C_r$ classes of real and $C_g$ classes of generated samples by maximizing $\bm{\lambda}$\footnote{Here it is unnecessary that $C_g = C_r$, e.g. training data contain positive and negative samples, for the task that only needs to generate positive samples, the negative samples can be employed to calibrate the gradients.}. And the generator enforces the distance to the decision hyper-plane of the desired class $\tilde{c}$ to be closer compared to all other decision hyper-planes by minimizing
\begin{equation}
    V(G) = \underset{\theta}{min} ~\underset{\bm{x}\sim P_r(\bm{x}), \bm{z}\sim P(\bm{z})}{\mathbb{E}} \sum_{c \in \{C_r, C_g\} } {\Big[ \mathcal{H}_c(\bm{u}) - \mathcal{H}_{\tilde{c}}(\bm{u}) \Big]}.
\end{equation}

Obviously, if $P_r(\bm{x}|y_c) = P_g(\bm{x}|y_c)$ for all $C$ classes, then the eigenvalues obtained by discriminating $C_r$ will be the same as that obtained by discriminating $\{C_r, C_g\}$. Thus we conjecture that, given a generator with infinite capacity, by gradually matching the mean of hidden representations $\bm{\mu}_c$ towards the desired class mean $\bm{\mu}_{\tilde{c}}$\footnote{More precisely, the similarity (inner product) between the projected hidden representation and class mean is increased.}, a Nash equilibrium of the generator and discriminator $(\theta^*, \phi^*)$ can be reached, where $\phi^*$ is the parameter of an \textit{optimal} classifier trained on $P_r(\bm{x}|y_c)$ (an empirical result is provided in \Cref{fig4:b}). In practice, since we do not have such information about the classifier during training of LD-GAN, the $\bm{\lambda}$ cannot be considered as a direct metric of divergence between $P_r(\bm{x})$ and $P_g(\tilde{\bm{x}})$. To avoid confusion, we employ a fixed update scheme rather than dynamic balancing in this paper for conditional generation experiments.

\begin{figure}[t]
\centering

    \begin{subfigure}{1\textwidth}
    \centering
        \begin{minipage}[c]{0.38\textwidth}
            \includegraphics[width=1\textwidth]{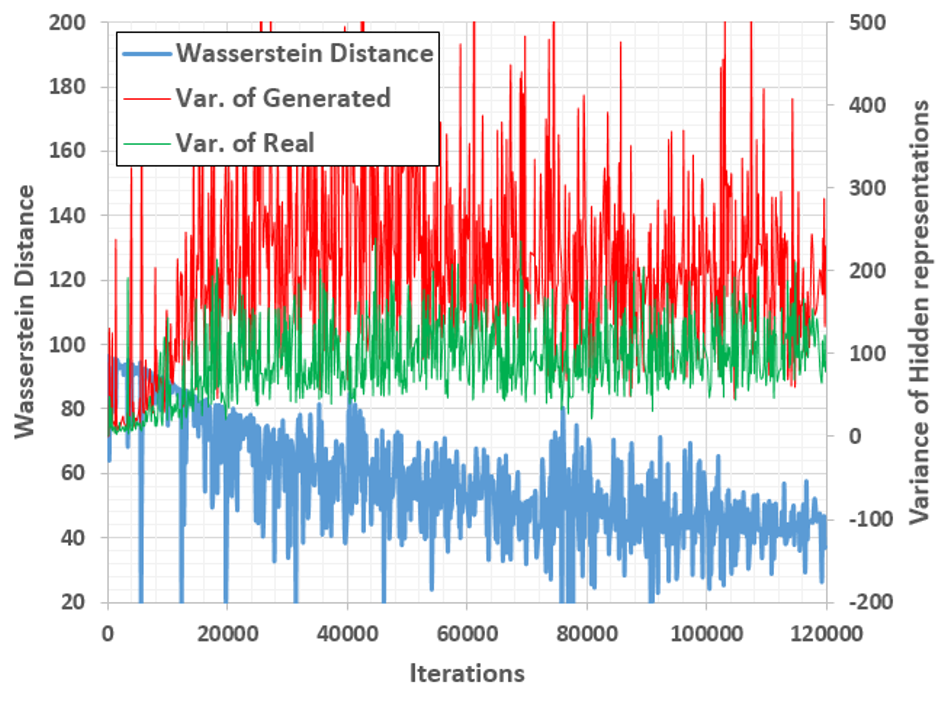}
        \end{minipage}
        \begin{minipage}[c]{0.06\textwidth}
            \crule[white]{0.3cm}{0.3cm}  
        \end{minipage}
        \begin{minipage}[c]{0.38\textwidth}
            \includegraphics[width=1\textwidth]{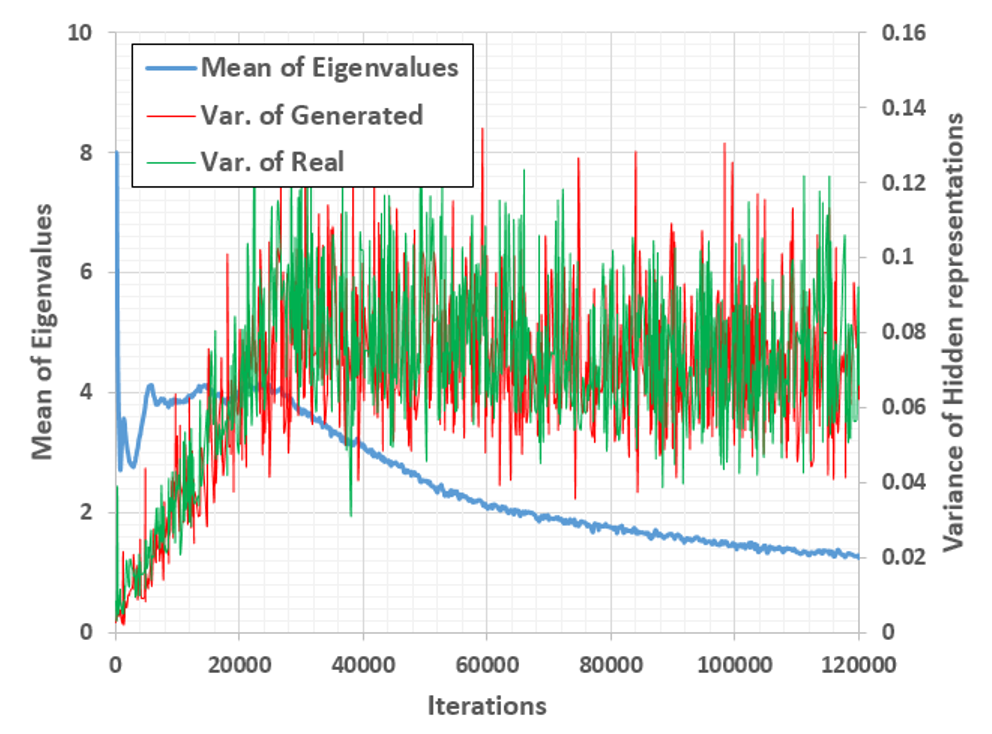}
        \end{minipage} \\
        \begin{minipage}[c]{0.96\textwidth}
            \centering
            \subcaption{Results obtained using weight clip.}
            \label{fig2:a}
        \end{minipage}
    \end{subfigure}
    
    \begin{subfigure}{1\textwidth}
    \centering
        \begin{minipage}[c]{0.38\textwidth}
            \includegraphics[width=1\textwidth]{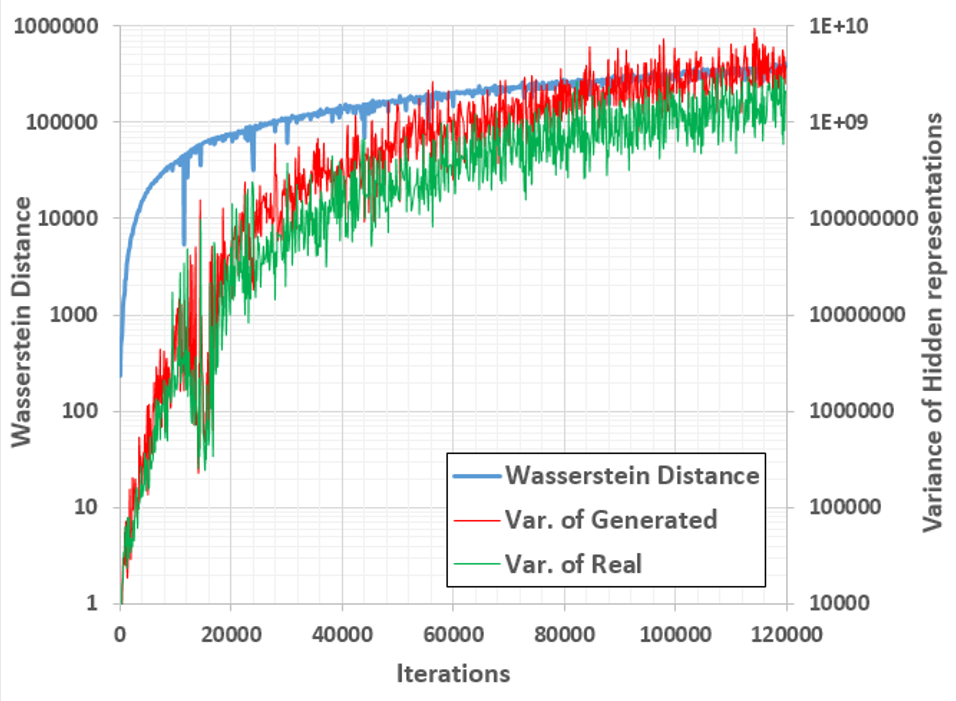}
        \end{minipage}
        \begin{minipage}[c]{0.06\textwidth}
            \crule[white]{0.3cm}{0.3cm}  
        \end{minipage}
        \begin{minipage}[c]{0.38\textwidth}
            \includegraphics[width=1\textwidth]{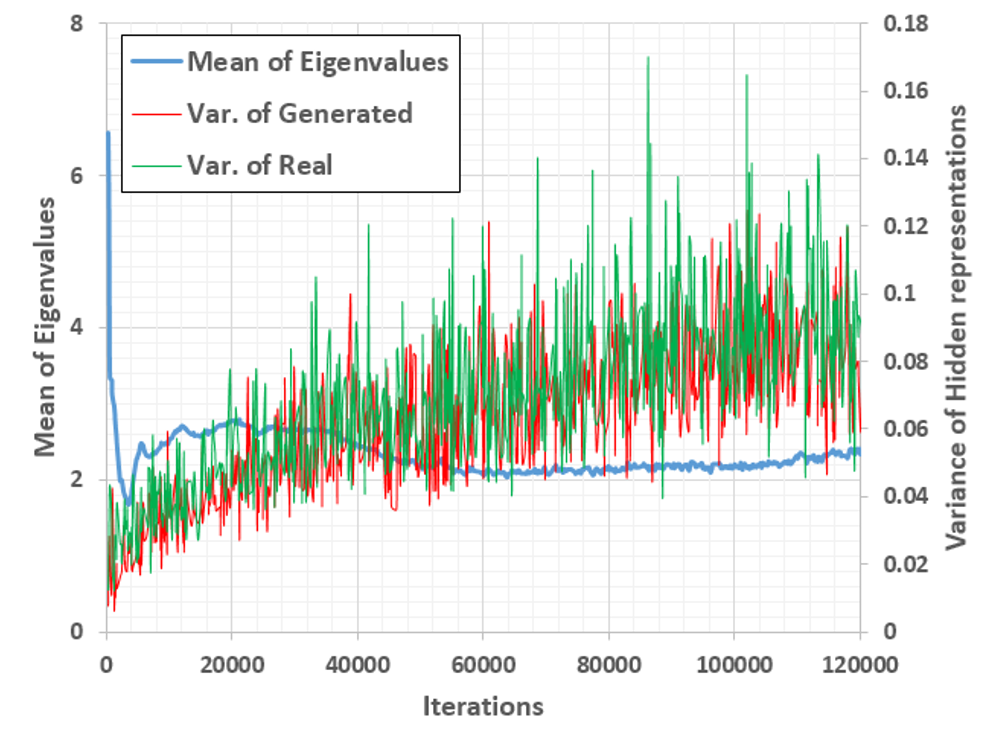}
        \end{minipage} \\
        \begin{minipage}[c]{0.96\textwidth}
            \centering
            \subcaption{Results obtained without using weight clip.}
            \label{fig2:b}
        \end{minipage}
    \end{subfigure}
    
    \caption{Comparison of variance of hidden representations of generated and real samples between WGAN and LD-GAN (a) with/(b) without using weight clip constraints. Convergence metrics are depicted with blue line (approximated Wasserstein distance and mean of eigenvalues for WGAN and LD-GAN, respectively). \textit{left}: WGAN, \textit{right}: LD-GAN}
    \label{fig2}
\end{figure}

\section{Experiments}
In this section, we experimentally analyze our proposed method in unsupervised and conditional generation tasks. For unsupervised generation, we use the bedroom subset of the LSUN dataset~(\citealt{yu2015lsun}), and for conditional generation, we use the MNIST dataset~(\citealt{lecun1998gradient}), and the whole LSUN dataset with all the 10 classes. The neural networks employed in the experiments are given in supplemental material.

\subsection{Stability analysis for unsupervised generation}
In this section, we analyze the stability of proposed methods using various configurations. First, we investigate the behaviour of WGAN and LD-GAN, using (\Cref{fig1:a}) and without using (\Cref{fig1:b}) weight clip. Then we setup two difficult configurations with which the WGAN is considered to be unstable, that is, no employment of BN in neither generator (G) nor discriminator (D) and optimization with Adam. Additionally, we remain the weight clip for training of WGAN, while the LD-GAN is trained without constrain on weights due to the unnecessity. The updated frequency for G and D are set to be $ln(\lambda)$ and $ln(1/\lambda)$ with a minimum $1$ per iteration, respectively. The results are provided in \Cref{fig1}, the proposed LD-GAN is able to generate authentic images in both standard and difficult configurations, with an unbounded objective for discriminator as aforementioned.

The variation of statistical moments observed during training is provided in \Cref{fig2}. It can be seen that, for standard WGAN trained with weight clip, after the discriminator reaches its upper bound of capacity (approximately 20K$\sim$30K iterations, when the variance of real samples stops growing), the mean and variance of real and fake samples start to get closer slowly. Once the constraints are removed, the divergence between mean of real and generated samples grows exponentially. Interestingly, the system is still able to generate authentic (but with low quality) images thanks to the exponential growth of variance. It is notable that, the variance of generated samples is remarkably larger than that of real samples, which suggests the difficulty of reducing divergence between distributions using the Wasserstein distance objective, and the quality of generated samples could be biased since some samples maybe distant from the true distribution even the mean discrepancy is small. On the other hand, the proposed LD-GAN shows better stability and consistency in variance between two distributions in both cases.

\begin{figure}[t]
\centering
    \begin{subfigure}[c]{0.37\textwidth}
        \includegraphics[width=1\textwidth]{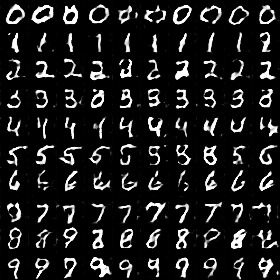}
        \subcaption{AC-WGAN}
        \label{fig3:a}
    \end{subfigure} 
    \begin{subfigure}[c]{0.06\textwidth}
        \crule[white]{0.4cm}{0.4cm}
    \end{subfigure}
    \begin{subfigure}[c]{0.37\textwidth}
        \includegraphics[width=1\textwidth]{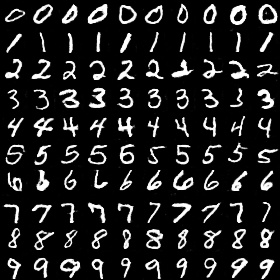}
        \subcaption{LD-GAN}
        \label{fig3:b}
    \end{subfigure} 
    \caption{Conditional generation using the MNIST digits.}
    \vspace{-10pt}
    \label{fig3}
\end{figure}

\begin{figure}[t]
\centering
    \begin{subfigure}[c]{0.37\textwidth}
        \includegraphics[width=1\textwidth]{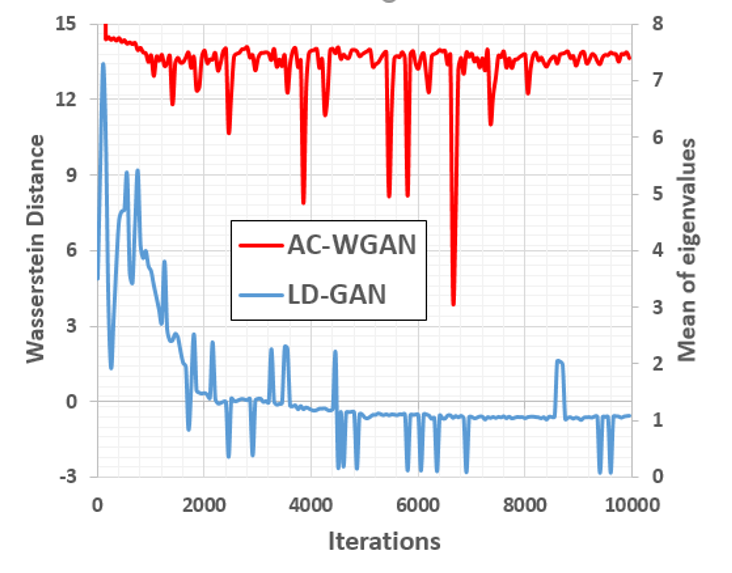}
        \subcaption{}
        \label{fig4:a}
    \end{subfigure} 
    \begin{subfigure}[c]{0.06\textwidth}
        \crule[white]{0.4cm}{0.4cm}
    \end{subfigure}
    \begin{subfigure}[c]{0.37\textwidth}
        \includegraphics[width=1\textwidth]{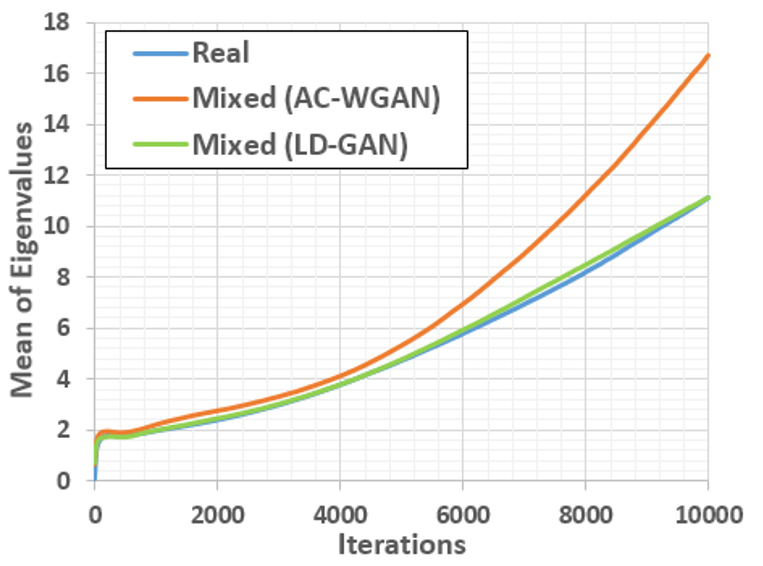}
        \subcaption{}
        \label{fig4:b}
    \end{subfigure}
    \vspace{-2pt}
    
    \caption{(a) Comparison of convergence using AC-WGAN and LD-GAN. The Wasserstein distance approximated by WGAN is given by red line, and the mean of eigenvalues obtained from training LDA is depicted by blue line. (b) Training process of a classifier on the dataset that only contains true samples (blue) and the dataset contains a mixture of true and generated samples (red and green).}
    \vspace{-10pt}
    \label{fig4}
\end{figure}

\subsection{Class Conditional generation}
In this section, we give the results for conditional generation using the proposed LD-GAN. For comparison, we employ a WGAN with an auxiliary classifier (\citealt{odena2016conditional}) using $1:1$ generation and classification loss as a reference model (denoted by AC-WGAN). We train an AC-WGAN model and a LD-GAN model on MNIST using a similar structure as utilized for generation of the bedroom dataset with minor modifications. The LD-GAN is implemented without imposing constraints on weights, and we employ RMSProp for optimization of both models. At each iteration, we update the discriminator 5 times and generator 1 time for training of AC-WGAN, and we update both discriminator and generator 2 times for training of LD-GAN. 

The images generated on the MNIST are presented in \Cref{fig3} in comparison with the results obtained using AC-WGAN. The results show that the digits generated using LD-GAN have better quality and diversity. In order to investigate the results, we further provide convergence results for both methods in \Cref{fig4:a}. It is observed that the mean of eigenvalues (blue line) decreases as training proceeds, and finally reaches to a stable range. However, we observe that the Wasserstein distance approximated by WGAN (red line) is almost left unchanged from the beginning of training. Therefore, we argue that, in AC-WGAN, it can be difficult to give consideration to both Wasserstein distance and classification loss, and the categorical cross-entropy loss obtained from classifiers contributed more in generating authentic digits, rather than the Wasserstein distance.

We further design an experiment to demonstrate the generalization of our proposed LD-GAN. We train two deep LDA classifiers (using the same architecture of the discriminator employed above) to classify 10 classes of real samples, and 10 classes of real in addition to 10 classes of generated samples (20 classes in total), respectively. According to the argument made in Section~\ref{subsec_condition}, both classifiers should provide similar eigenvalues since they cannot separate the real and generated samples belonging to the same class efficiently. Otherwise, the eigenvalues should be different due to a change in $\bm{S_b}$ when real and generated samples can be separated. In the results given in \Cref{fig4:b}, the \textit{Mixed} stands for training a classifier with mixed real and generated samples. It can be seen that, the increase of mean of eigenvalues shows high consistency between classifiers trained with real and generated samples from LD-GAN. We also provide the results obtained using samples generated with AC-WGAN for reference, the larger mean of eigenvalues in this case indicates a better separation capacity of the classifier for real and generated samples, as observed in \Cref{fig3:a}. 

\begin{figure}[t]
\centering
    \begin{subfigure}[c]{0.40\textwidth}
        \includegraphics[width=1\textwidth]{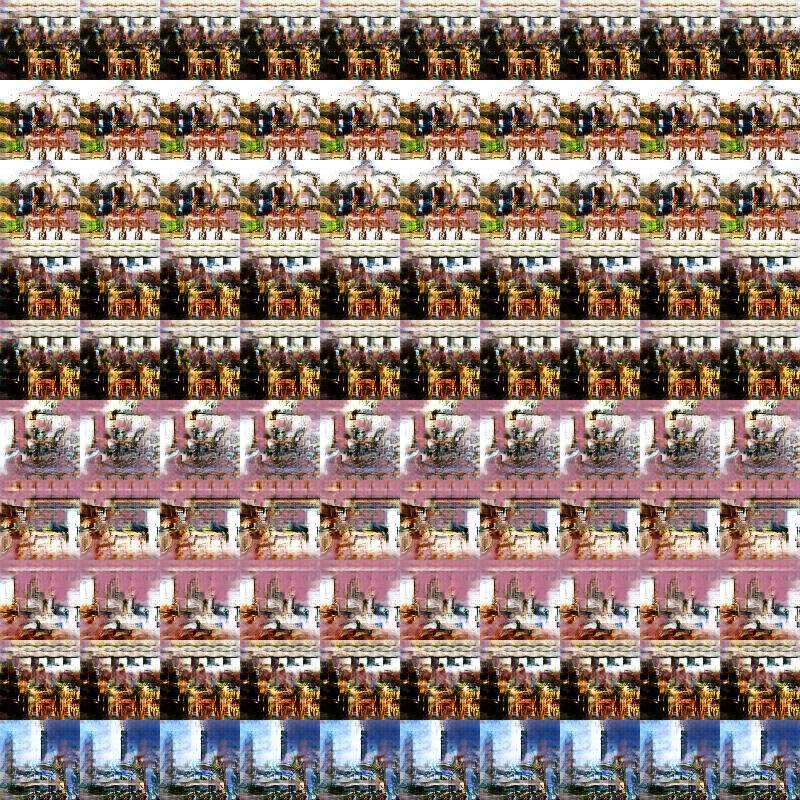}
        \subcaption{AC-WGAN}
        \label{fig5:a}
    \end{subfigure} 
    \begin{subfigure}[c]{0.07\textwidth}
        \crule[white]{0.7cm}{0.7cm}
    \end{subfigure}
    \begin{subfigure}[c]{0.40\textwidth}
        \includegraphics[width=1\textwidth]{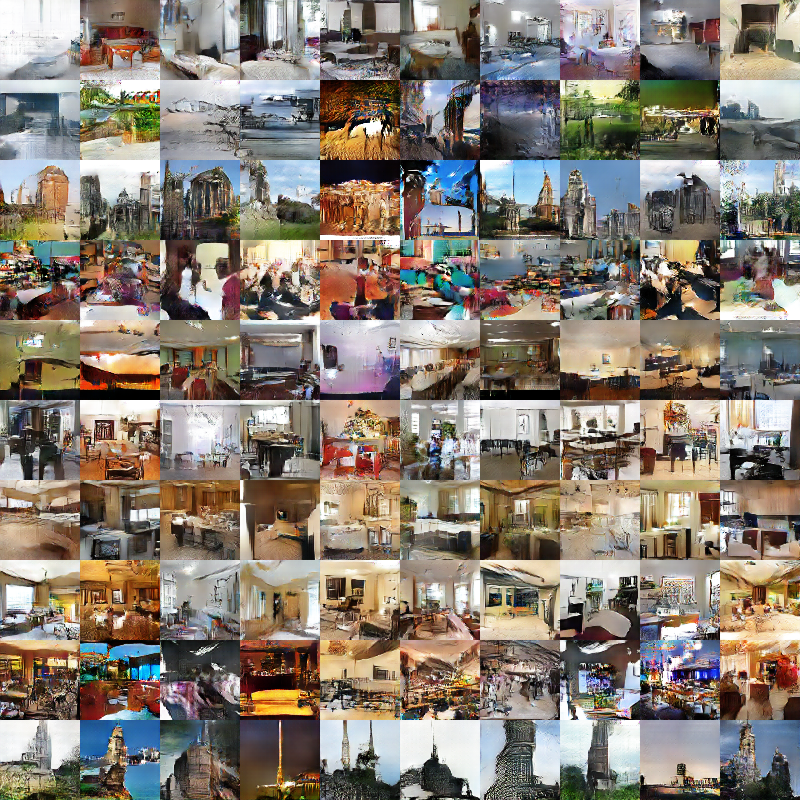}
        \subcaption{LD-GAN}
        \label{fig5:b}
    \end{subfigure} 
    
    \caption{Conditional generation with 10 classes of LSUN.}
    \label{fig5}
    \vspace{-10pt}
\end{figure}

In order to demonstrate the stability of the proposed method, we further employ the whole LSUN (using all 10 classes) datasets, and train a pair of generator and discriminator which are 3 and 2 layers deeper than formerly used for the bedroom dataset using both models, thus the classifiers trained using cross-entropy loss in AC-WGAN can barely contribute to the generation of authentic images. The results given in Figure~\ref{fig5} show that, AC-WGAN model fails to generate meaningful images and collapses to identical patterns for each class, while the proposed LD-GAN was still able to generate scene images with reasonable quality.

\begin{table}[h]
  \centering
  \caption{Neural network configurations for LSUN datasets. The DeConvolution and Convolution layer parameters are denoted by (De)Conv --\textless kernel size-stride\textgreater--number of output channels--activation and normalization. The fully connected layer parameters are denoted by Fc-number of output features, no activation nor batch normalization is employed after Fc layers. For conditional generation of LSUN full datasets, the outputs of AC-WGAN are Wasserstein distance and scores of 10 classes. For LD-GAN, the discriminator outputs 64 features to LDA. \textit{top}: generators, \textit{bottom}: discriminators.}
  \label{table1}
  \begin{tabular}{cc}
    \toprule
    \textbf{LSUN Bedroom} & \textbf{LSUN full} \\
    \midrule
    DeConv--\textless 4$\times$4-1$\times$1\textgreater--1024--LeakyRelu+BN & DeConv--\textless 5$\times$5-1$\times$1\textgreater--512--LeakyRelu+BN\\
    DeConv--\textless 4$\times$4-2$\times$2\textgreater--512--Relu+BN & DeConv--\textless 3$\times$3-2$\times$2\textgreater--512--Relu+BN\\
    DeConv--\textless 4$\times$4-2$\times$2\textgreater--256--Relu+BN & DeConv--\textless 3$\times$3-1$\times$1\textgreater--512--Relu+BN\\
    DeConv--\textless 4$\times$4-2$\times$2\textgreater--128--Relu+BN & DeConv--\textless 3$\times$3-2$\times$2\textgreater--384--Relu+BN\\
    DeConv--\textless 4$\times$4-2$\times$2\textgreater--3--Tanh & DeConv--\textless 3$\times$3-1$\times$1\textgreater--384--Relu+BN\\
    & DeConv--\textless 3$\times$3-2$\times$2\textgreater--256--Relu+BN\\
    & DeConv--\textless 3$\times$3-2$\times$2\textgreater--128--Relu+BN\\
    & DeConv--\textless 3$\times$3-1$\times$1\textgreater--3--Tanh\\
    output size: 64$\times$64 & output size: 80$\times$80 \\
    \midrule
    Conv--\textless 4$\times$4-2$\times$2\textgreater--64--LeakyRelu & Conv--\textless 3$\times$3-2$\times$2\textgreater--64--LeakyRelu\\
    Conv--\textless 4$\times$4-2$\times$2\textgreater--128--LeakyRelu+BN & Conv--\textless 3$\times$3-2$\times$2\textgreater--128--LeakyRelu+BN\\
    Conv--\textless 4$\times$4-2$\times$2\textgreater--256--LeakyRelu+BN & Conv--\textless 3$\times$3-2$\times$2\textgreater--256--LeakyRelu+BN\\
    Conv--\textless 4$\times$4-2$\times$2\textgreater--512--LeakyRelu+BN & Conv--\textless 3$\times$3-1$\times$1\textgreater--256--LeakyRelu+BN\\
     & Conv--\textless 3$\times$3-2$\times$2\textgreater--512--LeakyRelu+BN\\
     & Conv--\textless 3$\times$3-2$\times$2\textgreater--512--LeakyRelu+BN\\
    Fc--1 & Fc--1+10(AC-WGAN)~/~64(LD-GAN)\\
    \bottomrule
  \end{tabular}
\end{table}

\section{Conclusions}
We introduce a novel approach to improve the stability and generalization performance of vanilla GAN for both unsupervised and class conditional generation. The proposed method employ Linear Discriminant Analysis on the top of the discriminator to maximize the linear separability between the distributions of generated and targeted samples. We experimentally show that the proposed LD-GAN is able to overcome the instability caused by the moment matching objective, and generate authentic images in both unsupervised and class conditional generation tasks.

\clearpage

\newpage

\small
\bibliography{nips_2017}

\begin{thebibliography}{}

\bibitem[Arjovsky and Bottou, 2017]{arjovsky2017towards}
Arjovsky, M. and Bottou, L. (2017).
\newblock Towards principled methods for training generative adversarial
  networks.
\newblock In {\em NIPS 2016 Workshop on Adversarial Training. In review for
  ICLR}, volume 2016.

\bibitem[Arjovsky et~al., 2017]{arjovsky2017wasserstein}
Arjovsky, M., Chintala, S., and Bottou, L. (2017).
\newblock Wasserstein gan.
\newblock {\em arXiv preprint arXiv:1701.07875}.

\bibitem[Bishop, 2006]{bishop:2006:PRML}
Bishop, C.~M. (2006).
\newblock {\em Pattern Recognition and Machine Learning}.
\newblock Springer.

\bibitem[Chen et~al., 2016]{ChenDHSSA16}
Chen, X., Duan, Y., Houthooft, R., Schulman, J., Sutskever, I., and Abbeel, P.
  (2016).
\newblock Infogan: Interpretable representation learning by information
  maximizing generative adversarial nets.
\newblock {\em CoRR}, abs/1606.03657.

\bibitem[Dorfer et~al., 2015]{dorfer2015deep}
Dorfer, M., Kelz, R., and Widmer, G. (2015).
\newblock Deep linear discriminant analysis.
\newblock {\em arXiv preprint arXiv:1511.04707}.

\bibitem[Dziugaite et~al., 2015]{dziugaite2015training}
Dziugaite, G.~K., Roy, D.~M., and Ghahramani, Z. (2015).
\newblock Training generative neural networks via maximum mean discrepancy
  optimization.
\newblock {\em arXiv preprint arXiv:1505.03906}.

\bibitem[Goodfellow et~al., 2014]{goodfellow2014generative}
Goodfellow, I., Pouget-Abadie, J., Mirza, M., Xu, B., Warde-Farley, D., Ozair,
  S., Courville, A., and Bengio, Y. (2014).
\newblock Generative adversarial nets.
\newblock In {\em Advances in neural information processing systems}, pages
  2672--2680.

\bibitem[Huang et~al., 2016]{huang2016stacked}
Huang, X., Li, Y., Poursaeed, O., Hopcroft, J., and Belongie, S. (2016).
\newblock Stacked generative adversarial networks.
\newblock {\em arXiv preprint arXiv:1612.04357}.

\bibitem[Ioffe and Szegedy, 2015]{ioffe2015batch}
Ioffe, S. and Szegedy, C. (2015).
\newblock Batch normalization: Accelerating deep network training by reducing
  internal covariate shift.
\newblock {\em arXiv preprint arXiv:1502.03167}.

\bibitem[Kingma and Ba, 2014]{kingma2014adam}
Kingma, D. and Ba, J. (2014).
\newblock Adam: A method for stochastic optimization.
\newblock {\em arXiv preprint arXiv:1412.6980}.

\bibitem[LeCun et~al., 1998]{lecun1998gradient}
LeCun, Y., Bottou, L., Bengio, Y., and Haffner, P. (1998).
\newblock Gradient-based learning applied to document recognition.
\newblock {\em Proceedings of the IEEE}, 86(11):2278--2324.

\bibitem[Ledig et~al., 2016]{ledig2016photo}
Ledig, C., Theis, L., Husz{\'a}r, F., Caballero, J., Cunningham, A., Acosta,
  A., Aitken, A., Tejani, A., Totz, J., Wang, Z., et~al. (2016).
\newblock Photo-realistic single image super-resolution using a generative
  adversarial network.
\newblock {\em arXiv preprint arXiv:1609.04802}.

\bibitem[Li et~al., 2015]{li2015generative}
Li, Y., Swersky, K., and Zemel, R. (2015).
\newblock Generative moment matching networks.
\newblock In {\em Proceedings of the 32nd International Conference on Machine
  Learning (ICML-15)}, pages 1718--1727.

\bibitem[{Mao} et~al., 2016]{2016arXiv161104076M}
{Mao}, X., {Li}, Q., {Xie}, H., {Lau}, R.~Y.~K., {Wang}, Z., and {Smolley},
  S.~P. (2016).
\newblock {Least Squares Generative Adversarial Networks}.
\newblock {\em ArXiv e-prints}.

\bibitem[Mathieu et~al., 2015]{mathieu2015deep}
Mathieu, M., Couprie, C., and LeCun, Y. (2015).
\newblock Deep multi-scale video prediction beyond mean square error.
\newblock {\em arXiv preprint arXiv:1511.05440}.

\bibitem[Metz et~al., 2016]{metz2016unrolled}
Metz, L., Poole, B., Pfau, D., and Sohl-Dickstein, J. (2016).
\newblock Unrolled generative adversarial networks.
\newblock {\em arXiv preprint arXiv:1611.02163}.

\bibitem[Mirza and Osindero, 2014]{mirza2014conditional}
Mirza, M. and Osindero, S. (2014).
\newblock Conditional generative adversarial nets.
\newblock {\em arXiv preprint arXiv:1411.1784}.

\bibitem[Mroueh et~al., 2017]{mroueh2017mcgan}
Mroueh, Y., Sercu, T., and Goel, V. (2017).
\newblock Mcgan: Mean and covariance feature matching gan.
\newblock {\em arXiv preprint arXiv:1702.08398}.

\bibitem[{Nowozin} et~al., 2016]{2016arXiv160600709N}
{Nowozin}, S., {Cseke}, B., and {Tomioka}, R. (2016).
\newblock {f-GAN: Training Generative Neural Samplers using Variational
  Divergence Minimization}.
\newblock {\em ArXiv e-prints}.

\bibitem[Odena et~al., 2016]{odena2016conditional}
Odena, A., Olah, C., and Shlens, J. (2016).
\newblock Conditional image synthesis with auxiliary classifier gans.
\newblock {\em arXiv preprint arXiv:1610.09585}.

\bibitem[Pang et~al., 2005]{pang2005incremental}
Pang, S., Ozawa, S., and Kasabov, N. (2005).
\newblock Incremental linear discriminant analysis for classification of data
  streams.
\newblock {\em IEEE Transactions on Systems, Man, and Cybernetics, Part B
  (Cybernetics)}, 35(5):905--914.

\bibitem[Poole et~al., 2016]{poole2016improved}
Poole, B., Alemi, A.~A., Sohl-Dickstein, J., and Angelova, A. (2016).
\newblock Improved generator objectives for gans.
\newblock {\em arXiv preprint arXiv:1612.02780}.

\bibitem[{Qi}, 2017]{2017arXiv170106264Q}
{Qi}, G.-J. (2017).
\newblock {Loss-Sensitive Generative Adversarial Networks on Lipschitz
  Densities}.
\newblock {\em ArXiv e-prints}.

\bibitem[Reed et~al., 2016]{reed2016generative}
Reed, S., Akata, Z., Yan, X., Logeswaran, L., Schiele, B., and Lee, H. (2016).
\newblock Generative adversarial text to image synthesis.
\newblock In {\em Proceedings of The 33rd International Conference on Machine
  Learning}, volume~3.

\bibitem[Salimans et~al., 2016]{SalimansGZCRC16}
Salimans, T., Goodfellow, I.~J., Zaremba, W., Cheung, V., Radford, A., and
  Chen, X. (2016).
\newblock Improved techniques for training gans.
\newblock {\em CoRR}, abs/1606.03498.

\bibitem[Salimans and Kingma, 2016]{salimans2016weight}
Salimans, T. and Kingma, D.~P. (2016).
\newblock Weight normalization: A simple reparameterization to accelerate
  training of deep neural networks.
\newblock In {\em Advances in Neural Information Processing Systems}, pages
  901--901.

\bibitem[Springenberg, 2015]{springenberg2015unsupervised}
Springenberg, J.~T. (2015).
\newblock Unsupervised and semi-supervised learning with categorical generative
  adversarial networks.
\newblock {\em arXiv preprint arXiv:1511.06390}.

\bibitem[Stuhlsatz et~al., 2012]{stuhlsatz2012feature}
Stuhlsatz, A., Lippel, J., and Zielke, T. (2012).
\newblock Feature extraction with deep neural networks by a generalized
  discriminant analysis.
\newblock {\em IEEE transactions on neural networks and learning systems},
  23(4):596--608.

\bibitem[Yan et~al., 2016]{yan2016attribute2image}
Yan, X., Yang, J., Sohn, K., and Lee, H. (2016).
\newblock Attribute2image: Conditional image generation from visual attributes.
\newblock In {\em European Conference on Computer Vision}, pages 776--791.
  Springer.

\bibitem[Yu et~al., 2015]{yu2015lsun}
Yu, F., Seff, A., Zhang, Y., Song, S., Funkhouser, T., and Xiao, J. (2015).
\newblock Lsun: Construction of a large-scale image dataset using deep learning
  with humans in the loop.
\newblock {\em arXiv preprint arXiv:1506.03365}.

\bibitem[Zhao et~al., 2016]{zhao2016energy}
Zhao, J., Mathieu, M., and LeCun, Y. (2016).
\newblock Energy-based generative adversarial network.
\newblock {\em arXiv preprint arXiv:1609.03126}.

\bibitem[Zhu et~al., 2016]{zhu2016generative}
Zhu, J.-Y., Kr{\"a}henb{\"u}hl, P., Shechtman, E., and Efros, A.~A. (2016).
\newblock Generative visual manipulation on the natural image manifold.
\newblock In {\em European Conference on Computer Vision}, pages 597--613.
  Springer.

\end{thebibliography}

\end{document}